\def\year{2018}\relax
\documentclass[letterpaper]{article} 
\usepackage{aaai18}  
\usepackage{amsmath}
\usepackage{times}  
\usepackage{helvet}  
\usepackage{courier}  
\usepackage{url}  
\usepackage{graphicx}  
\usepackage{amssymb}
\newcommand{\eg}{e.g.}
\newcommand{\ie}{i.e.}
\begin{document}
	%
	\title{r-BTN: Cross-domain Face Composite and Synthesis from Limited Facial Patches}
	\author{Yang Song, Zhifei Zhang, Hairong Qi\\
		Department of Electrical Engineering and Computer Science\\
		University of Tennessee, Knoxville, TN 37996, USA\\
		{\{ysong18, zzhang61, hqi@utk.edu\}}\\
	}
	\maketitle
\begin{abstract}
Recent face composite and synthesis related works have shown promising results in generating realistic face images from deep convolutional networks. 
However, these works either do not generate consistent results when the constituent patches contain large domain variations (\ie, from face and sketch domains) or cannot generate high-resolution images with limited facial patches (\eg, the inpainting approach tends to blur the generated region when the missing area is more than 50\%). Motivated by the mental imagery and simulation in human cognition, we exploit the potential of deep learning networks in filling large missing region (\eg, as high as 95\% missing) and generating realistic faces with high-fidelity in cross domains. We propose the recursive generation by bidirectional transformation networks (r-BTN) that recursively generates a whole face/sketch from a small sketch/face patch. The large missing area and domain variations make it difficult to generate satisfactory results using a unidirectional cross-domain learning structure. We explore that the bidirectional transformation network can lead to the consistent result by minimizing the forward and backward errors in the cross-domain scenario. On the other hand, a forward and backward bidirectional learning between the face and sketch domains would enable recursive estimation of the missing region in an incremental manner to yield appealing results. r-BTN also adopts an adversarial constraint to encourage the generation of realistic faces/sketches.  Extensive experiments have been conducted to demonstrate the superior performance from r-BTN as compared to existing potential solutions. 
\end{abstract}
\section{Introduction}
\begin{figure}
		\centering
		\includegraphics[width=\columnwidth]{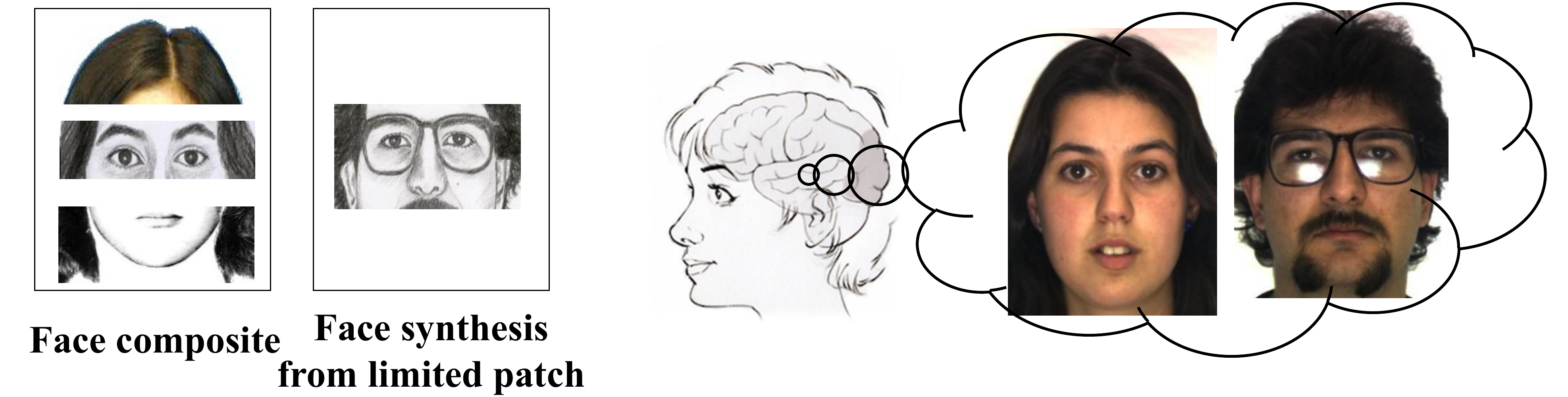}
		\caption{Illustration of face composite based on cross-domain patches and face synthesis from limited facial patch.}
		\label{fig:demo_task}
	\end{figure}

We start by asking an interesting yet challenging question, ``If provided with limited facial patches from sketch/face domains where human beings may be able to generate a real face image in brain~\cite{kosslyn2006case} as shown in Fig.~\ref{fig:demo_task}, can advanced computer vision techniques generate the whole face image?'' Recently, several face synthesis methods built on neural networks have emerged~\cite{zhang2017age,sangkloy2016scribbler}.
These methods can generate face/sketch images based on whole face information from one domain. However, how to generate realistic faces/sketches
that are consistent to the given sketch/face patches is still
a challenging task because large missing area could lead
to blurry generated images. In addition, some existing methods (\eg, Photofit~\cite{photofit}) synthesize faces by stitching patches
from cross domains which deteriorates the consistency
and photo-reality. It is still unclear how to preserve the color/domain consistency between patches with large domain variations.

In this paper, we study the above-mentioned problems which would play a key role in many applications, such as face image stitching, face blending, face editing, etc. To the best of our knowledge, this work represents the first attempt to cross-filling large missing area in both face and sketch domains. Existing works that may potentially address this problem are mainly in the perspectives of face/sketch synthesis/transformation and image inpainting. 
	
    The face/sketch synthesis works~\cite{wang2009face,tang2003face,zhou2012markov,song2014real} synthesize target faces from the source domain through patch-wise searching of similar patches in the training set. Without the generative capability, these methods fail to render reasonable pixels for large missing areas. The rapid development of generative adversarial networks (GANs)~\cite{goodfellow2014generative} has shown impressive performance in face generation~\cite{radford2015unsupervised,zhang2017age}, domain transformation~\cite{zhu2016generative,isola2016image}, and inpainting~\cite{yeh2016semantic,pathak2016context}. However, generating faces from small patches in either single or cross domains has not been explored. Intuitively, combining domain transformation and inpainting works could be a potential solution. However, with large missing area, the generated results tend to be blurred and may look unrealistic. 
    
\begin{figure*}[t]
		\centering
		\includegraphics[width=2\columnwidth]{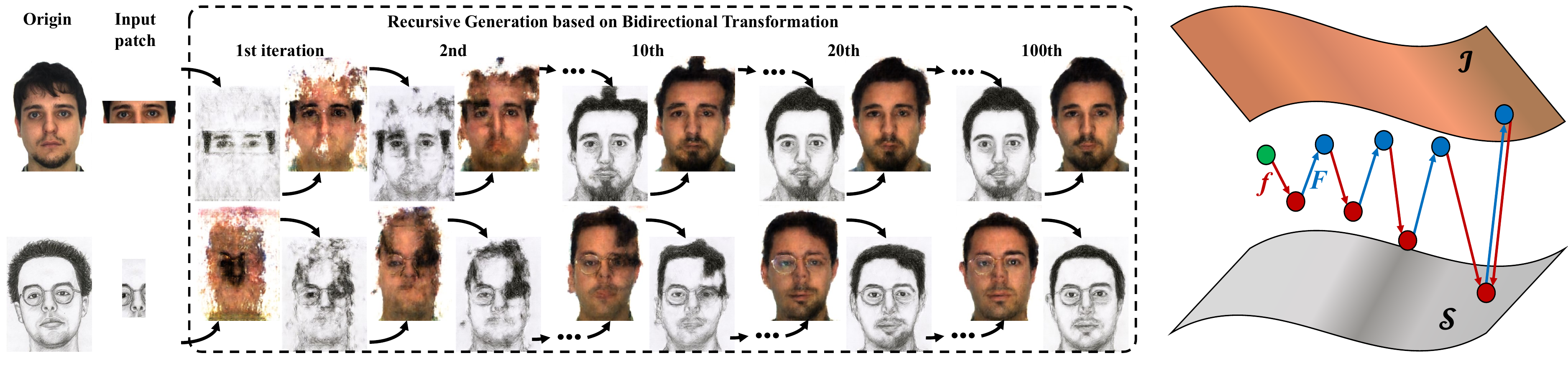}
		\caption{Examples of recursive generation from small patches by the bidirectional transformation network. Left: Original face/sketch and the corresponding input patches extracted from them. Inside of the dashed box demonstrates the generated face/sketch at different iteration steps. Right: Illustration of transformation between the face and sketch manifolds $\mathcal{I}$ and $\mathcal{S}$, respectively. The green dot denotes a given face patch. The red and blue arrows are the learned mapping $f$ and $F$, respectively. The red and blue dots are generated sketches and faces through corresponding mapping.}
		\label{fig:demo}
	\end{figure*}

	In this paper, we investigate the problem of cross-domain face/sketch generation conditioned on a given small patch of sketch/face. We assume that faces and sketches lie on high-dimensional manifolds $\mathcal{I}$ and $\mathcal{S}$, respectively, as shown in Fig.~\ref{fig:demo} (right). The given small sketch/face patch will initially deviate from the corresponding manifold due to large amount of missing data. With the learned bidirectional transformation network (BTN), \ie, $f$ and $F$, the given patch will be recursively mapped forward and backward between $\mathcal{I}$ and $\mathcal{S}$. Each mapping will yield a result progressively closing in onto either the face or sketch manifold, and eventually approaching the real whole face/sketch images as shown in Fig.~\ref{fig:demo} (middle). 
	An adversarial network is imposed on both $f$ and $F$, forcing more photo-realistic faces/sketches. The rationale and benefit of the proposed r-BTN will be further discussed.
	
	This paper makes the following contributions: 
	1) We tackle the challenging problem of face/sketch generation from small patches, estimating large missing area based on limited information while alleviating the blur effect suffered by existing works. 
	2) We propose the recursive generation by bidirectional transformation networks (r-BTN), which learns both a forward and backward mapping function between cross domains to enable a recursive update of the generated faces/sketches for more consistent and high-fidelity results even with large portions of missing data.
	3) We further exploit the capacity of r-BTN in fusing multiple patches from multiple domains and multiple people (\ie, face composite) to output a realistic and consistent face in a generative manner. 

	\section{Related Works}
	We will discuss related works from three closely related areas, namely, face/sketch synthesis/transformation, image inpainting, and face manipulation.
	
\noindent\textbf{Face/Sketch Synthesis/Transformation}
	related works mainly fall into two categories: matching-based and generation-based methods.
	Most face/sketch synthesis works~\cite{wang2009face,zhang2010lighting,zhou2012markov} are matching-based, which synthesize faces from best matched patches by searching from the training dataset. For example, \cite{wang2009face} divided a given face/sketch image into patches, each of which was matched to a series of similar patches from the training dataset. Then, the patches in the target domain corresponding to the matched patches were stitched via Markov random field to synthesize a transformed face. The matching-based methods have two drawbacks: 1) The matching procedure is time-consuming for a large training dataset, and 2) they cannot effectively estimate the patch content from missing area. 	
	The generation-based methods~\cite{taigman2016unsupervised,isola2016image} are mainly developed from encoder-decoder networks and adversarial generative networks. For example, \cite{isola2016image,zhu2017unpaired} proposed a general domain transformation method through conditional generative adversarial network. It could also be utilized for face/sketch transformation. However, it is not trained for the purpose of estimating missing areas. Moreover, to achieve bidirectional face/sketch transformation, two transformation networks (\ie, face to sketch and sketch to face) need to be learned independently.   
	
	\noindent\textbf{Image Inpainting}
	aims to fill in unwanted or missing part of an image. Most inpainting methods~\cite{efros1999texture,shen2002mathematical,criminisi2004region} estimate the missing part based on surrounding pixels, and therefore are not suitable for filling in large missing areas. Although some recent works~\cite{yeh2016semantic,pathak2016context} claimed the ability of filling in up to 80\% missing regions, 
	they tend to generate blurred results, which may be with visible inconsistency between the given and estimated areas. In addition, inpainting related methods train on randomly masked inputs and perform filling in a single domain, while the proposed work uses the whole face/sketch pairs in training and perform cross-domain filling.

	\noindent\textbf{Face Manipulation}
	works~\cite{zhang2017age,yan2016attribute2image} could be a potential solution to the proposed task because they can generate faces by manipulating the latent variables. Given a small patch, they may search the latent space for a best matched face. Thus, the generative model performs like matching-based methods which may be time-consuming. A more efficient way is to minimize the error between the generated face and the given patch.
	However, it cannot ensure consistent results because only the patch location (where the error comes from) will be updated regardless of its surroundings. 

	 
	\section{The Bidirectional Transformation Network}
	\label{sec:methodology}
	In this section, we first elaborate on the benefit of the proposed BTN through a comparison with unidirectional transformations. This is followed by a detailed description of the training and testing stages of the proposed r-BTN.
	The training stage learns the bidirectional transformation between the face and sketch domains using whole face/sketch pairs. The testing stage recursively generates the whole face/sketch from given small sketch/face patches.
	
	\subsection{The Bidirectional Network Structure}    
	\label{sec:btn}
\begin{figure}[h]
		\centering
		\includegraphics[width=0.9\columnwidth]{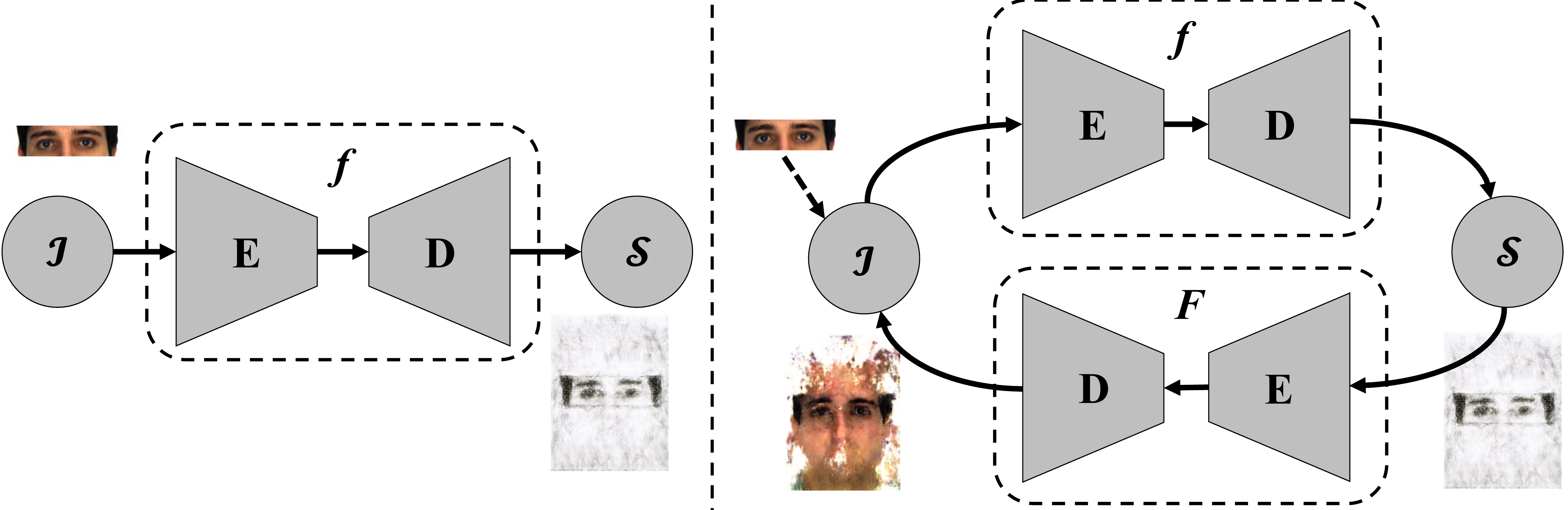}
		\caption{Comparison of unidirectional and bidirectional transformations between $\mathcal{I}$ and $\mathcal{S}$ domains. E and D are the encoder-decoder networks. The patch (eyes) generates the sketch, and then the sketch is transformed back where facial outline has been estimated.}
		\label{fig:univsbi}
	\end{figure}

	Assume a training set in $\mathcal{I}\times \mathcal{S}$, where $\mathcal{I}$ and $\mathcal{S}$ denote the face  and sketch  domains, respectively. The unidirectional transformation, \eg, \cite{isola2016image}, learns a mapping $f:\mathcal{I}\rightarrow\mathcal{S}$ which could be implemented by encoder-decoder networks, as shown in Fig.~\ref{fig:univsbi} (left). 
	The BTN, on the other hand, simultaneously involves the forward mapping $f$ and backward mapping $F:\mathcal{S}\rightarrow\mathcal{I}$, as shown in Fig.~\ref{fig:univsbi} (right). The bidirectional transformation forms a closed loop where the output of $f$ serves as the input to $F$, and the output of $F$ serves as  the input to $f$ in the next iteration.  The forward transformation $f$ may discard information in general due to the domain difference (\eg, color information will be discarded from $\mathcal{I}$ to $\mathcal{S}$), but the backward transformation $F$ closes the loop by connecting the output from $f$ in the $\mathcal{S}$ domain and the original input in the $\mathcal{I}$ domain and generates an intermediate result in $\mathcal{I}$ where additional face information (\eg, facial outline) has been estimated and the discarded information (\eg, color) restored. The bidirectional network structure enables the recursive update of the face (from $F$) and sketch (from $f$), taking advantage of the progressively learned knowledge in both domains and generate full face/sketch with high fidelity.
	
	The effectiveness of the recursive bidirectional transformation between face and sketch domains is well demonstrated in Fig.~\ref{fig:demo}. In general, the missing area is roughly filled at the beginning (iteration 1 and 2) although it is blurred. Then, facial details are progressively enhanced (iteration 10) and sharpened (iteration 20). Finally, a realistic face/sketch, including reasonable hair style, is generated. Because of the very limited information provided in the input patch, it is difficult to generate a face/sketch exactly the same as the original. However, the generated face/sketch still preserves the pixel-level content of the given patch. 
	
	
	\subsection{Training Stage} 
	\label{subsec:train}
\begin{figure}[h]
		\centering
		\includegraphics[width=0.8\columnwidth]{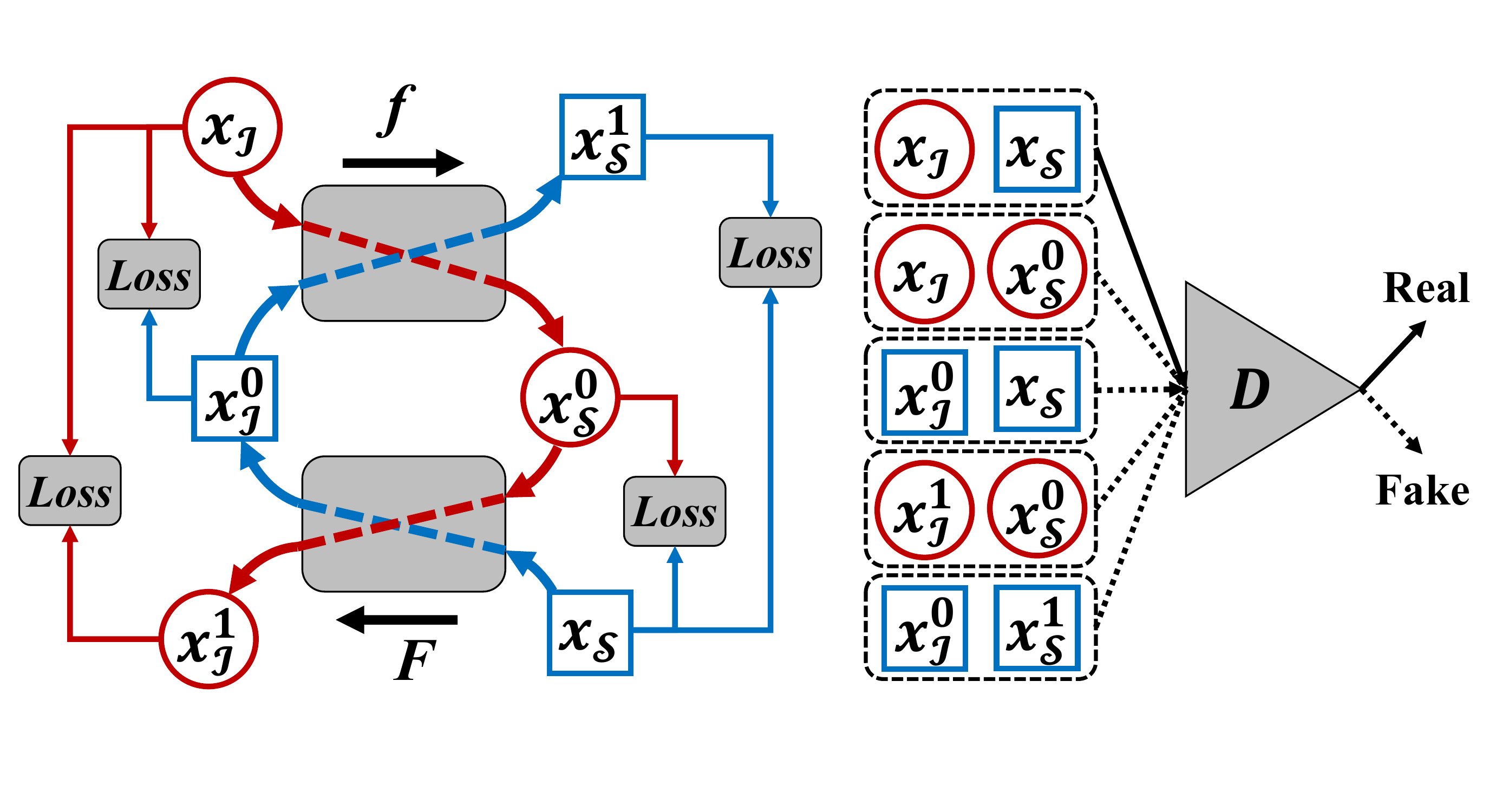}
		\caption{Training flow of the bidirectional transformation network. $x_\mathcal{I}$ and $x_\mathcal{S}$ are the real face/sketch pair. Red and blue arrows denote the transformation paths of $x_\mathcal{I}$ and $x_\mathcal{S}$, respectively. The transformation functions $f$ and $F$ could be encoder-decoder networks. $Loss$ denotes the $\ell_1$-norm. The discriminator $D$ is trained on real and generated (fake) face/sketch pairs.}
		\label{fig:flow}
	\end{figure}

	Fig.~\ref{fig:flow} illustrates the details of the BTN structure where the mapping functions, $f$ and $F$, are learned in a bidirectional fashion instead of the commonly used unidirectional mapping. 
	
	Given the original face/sketch pair $x_\mathcal{I}$ and $x_\mathcal{S}$, the following transformations are performed,
	\begin{align*}
		x_\mathcal{S}^0 = f(x_\mathcal{I}),\;  & x_\mathcal{I}^1 = F(x_\mathcal{S}^0)=F(f(x_\mathcal{I})),  \\
		x_\mathcal{I}^0 = F(x_\mathcal{S}),\;  & x_\mathcal{S}^1 =f(x_\mathcal{I}^0)=f(F(x_\mathcal{S})). \\ 
	\end{align*}
	The objective is to learn the bidirectional transformations between $\mathcal{I}$ and $\mathcal{S}$, so that any face/sketch pair could be uniquely mapped forward and backward into another domain. 
	To achieve invertible transformation, \ie, preserving the identity of face and sketch during transformations, we minimize the reconstruction error $\mathcal{L}_{rec}$ between real and generated faces or sketches as Eq.~\ref{eq:rec_err}. 
	\begin{equation}
		\mathcal{L}_{rec} = \sum_{i=0}^{1}\left(\| x_\mathcal{I} - x_\mathcal{I}^i\|_1 + \| x_\mathcal{S} - x_\mathcal{S}^i\|_1 \right),
		\label{eq:rec_err}
	\end{equation}   
	where the $\ell_1$-norm instead of the $\ell_2$-norm is used to avoid blurry results. 
	Besides $\mathcal{L}_{rec}$, an adversarial constraint is employed to encourage photo-realistic face/sketch pairs. The discrimination loss can be written as
	\begin{equation}
		\mathcal{L}_{adv} = \mathbb{E}_{\omega\in {\Omega}}\left[ \log D(\omega) \right] - \mathbb{E}_{\substack{x_\mathcal{I}\in \mathcal{I}\\x_\mathcal{S}\in \mathcal{S}}}[\log D(x_\mathcal{I}, x_\mathcal{S})],
		\label{eq:adv_err}
	\end{equation}
	where 
	\begin{align*}
\Omega=&\left\{ (x_\mathcal{I},x_\mathcal{S}^0)_j,  (x_\mathcal{I}^1,x_\mathcal{S}^0)_j, (x_\mathcal{I}^0,x_\mathcal{S})_j, (x_\mathcal{I}^0,x_\mathcal{S}^1)_j\right\} \\
=&\left\{ (x_\mathcal{I},f(x_\mathcal{I}))_j,  (F(f(x_\mathcal{I}),f(x_\mathcal{I})))_j,\right. \\
~ &\;\;\left. (F(x_\mathcal{S}),x_\mathcal{S})_j, (F(x_\mathcal{S}),f(F(x_\mathcal{S})))_j \right\}
\end{align*}
	indicates the fake face/sketch pairs, and $j$ indexes the fake pairs generated from the $j$th real pair in a mini-batch. Note that only $\left(  x_\mathcal{I}, x_\mathcal{S} \right)$ is the real pair. 
	Combining Eqs.~\ref{eq:rec_err} and \ref{eq:adv_err}, the objective function is 
	\begin{equation}
		\min_{f,F,D}\; \mathcal{L}_{adv} + \lambda\mathcal{L}_{rec},
		\label{eq:objective}
	\end{equation} 
	where $\lambda$ balances the adversarial loss and reconstruction loss.
	In optimization, $f$, $F$, and $D$ are updated alternatively. The discriminator $D$ is updated by minimizing $\mathcal{L}_{adv}$. The update of $f$ and $F$ is performed by 
\begin{align}
\min_{f} \; & \mathbb{E}_{\omega\in\Omega_f}\left[ \log D(\omega) \right] + \lambda\sum_{i=0}^{1} \| x_\mathcal{S} - x_\mathcal{S}^i\|_1 \label{eq:min_f}, \\
\min_{F} \; & \mathbb{E}_{\omega\in\Omega_F}\left[ \log D(\omega) \right] + \lambda\sum_{i=0}^{1} \| x_\mathcal{I} - x_\mathcal{I}^i\|_1 \label{eq:min_F}, 
\end{align}
	where 
\begin{align*}
\Omega_f=& \left\{(x_\mathcal{I},x_\mathcal{S}^0)_j, (x_\mathcal{I}^0,x_\mathcal{S}^1)_j\right\}\\
 =& \left\{(x_\mathcal{I},f(x_\mathcal{I}))_j, (x_\mathcal{I}^0,f(x_\mathcal{I}^0))_j  \right\},\\
\Omega_F=& \left\{(x_\mathcal{I}^0,x_\mathcal{S})_j, (x_\mathcal{I}^1,x_\mathcal{S}^0)_j\right\}\\
 =& \left\{(F(x_\mathcal{S}),x_\mathcal{S})_j, (F(x_\mathcal{S}^0),x_\mathcal{S}^0)_j\right\},
\end{align*}
	and $\Omega=\Omega_f \cup \Omega_F$. Here, $j$ is again the index of training samples in a mini-batch.
	
	\subsection{Testing Stage} 
	\label{subsec:test}
\begin{figure}[h]
		\centering
		\includegraphics[width=0.8\columnwidth]{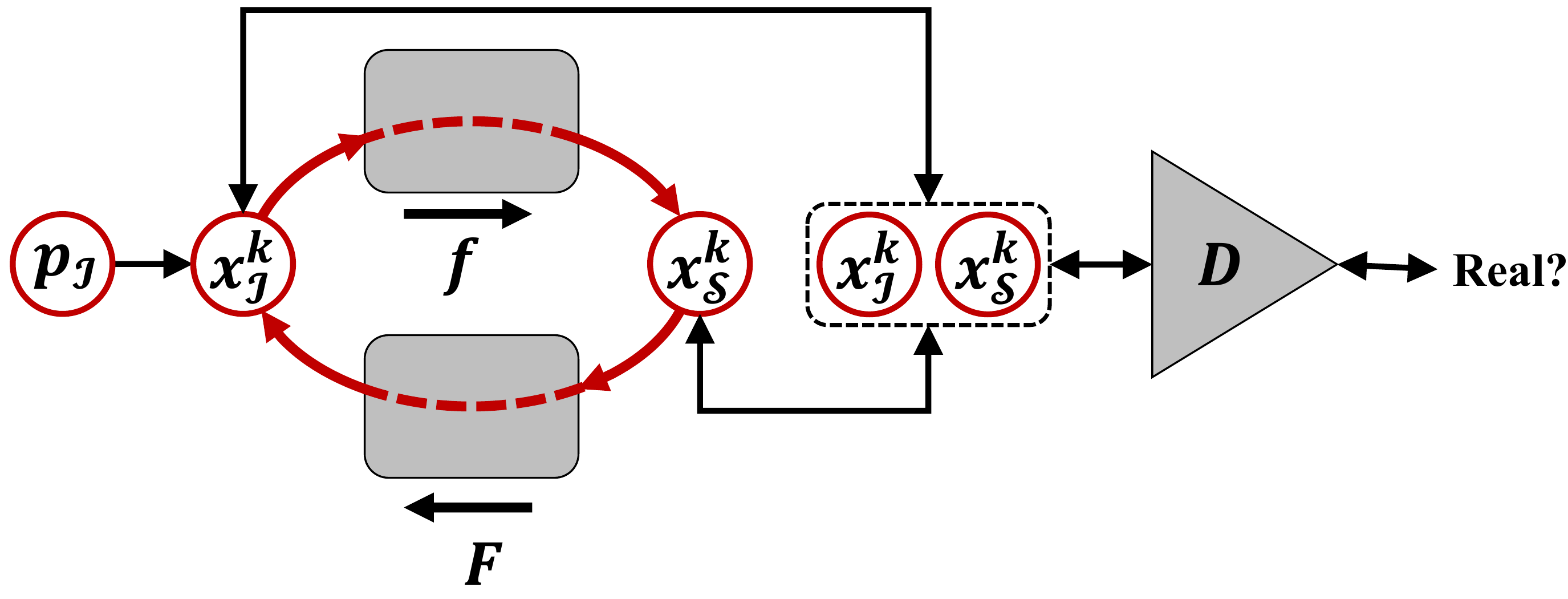}
		\caption{Testing flow of r-BTN, assuming a face patch $p_\mathcal{I}$ as the input. At step $k$, the generated face is $x_\mathcal{I}^k$. Replacing the corresponding area of $x_\mathcal{I}^k$ by the patch $p_\mathcal{I}$ and transforming $x_\mathcal{I}^k$ to $x_\mathcal{S}^k$, we get a face/sketch pair $\left(x_\mathcal{I}^k,x_\mathcal{S}^k\right)$. Then, this pair is adjusted by the error back propagated from $D$ as comparing to the output of real pairs. Finally, $x_\mathcal{S}^k$ is transformed back to the face domain, generating $x_\mathcal{I}^{k+1}$.}
		\label{fig:test_flow}
	\end{figure}

	During testing, given an arbitrary patch from either domain, a whole face from the other domain could be generated in a recursive manner through the bidirectional transformation. The testing flow is shown in Fig.~\ref{fig:test_flow}, 
	which demonstrates the case of given a face patch $p_\mathcal{I}$. Similarly, if a sketch patch $p_\mathcal{S}$ is given, it will be fed to $x_{\mathcal{S}}$ and similar testing flow can be carried out to generate a whole face image.
	In this paper, a patch is created through multiplying a whole face/sketch by a mask $M$, \eg, $p_\mathcal{I} = x_{\mathcal{I}} \odot M$ where $\odot$ denotes the element-wise multiplication. 
	
	The bidirectional transformation network structure enables a recursive generation between sketches and faces. 
	Given the current result $x_\mathcal{I}^k$, the next generation $x_\mathcal{I}^{k+1}$ can be obtained by
	\begin{align}
		{x}_\mathcal{I}^{k} &\leftarrow x_\mathcal{I}^{k} \odot (1-M) + p_\mathcal{I}  \label{eq:cast},\\
		{x}_\mathcal{S}^{k} &\leftarrow f\left({x}_\mathcal{I}^{k}\right) \label{eq:I2S},\\
		{x}_\mathcal{S}^{k} &\leftarrow {x}_\mathcal{S}^{k} - \frac{\partial D(x_\mathcal{I}^{k}, x_\mathcal{S}^{k})}{\partial x_\mathcal{S}^{k}} \label{eq:adjust},\\
		{x}_\mathcal{I}^{k+1} &\leftarrow F\left( {x}_\mathcal{S}^{k} \right).
		\label{eq:S2I}
	\end{align} 
	
	In order to generate photo-realistic faces/sketches such that the given patch and the estimated complement blend together in a consistent fashion, we have applied two constraints during the recursive generation process. First, we keep the given patch, $p_\mathcal{I}$, as the anchor that remains the same across different iterations. In other words, $p_\mathcal{I}$ directly covers the corresponding area of the newly generated face to explicitly preserve the given content (Eq.~\ref{eq:cast}). Then, $x_\mathcal{I}^k$ is transformed to the sketch domain by $f$ (Eq.~\ref{eq:I2S}). 
	Unlike most GANs related works which utilize $D$ only in the training stage, we utilize $D$ as a second constraint in the testing process to ensure realistic faces/sketches generation in each iteration such that small deviations get to be corrected instead of accumulated through iterations. 
	
	Given a small patch, the testing stage needs multiple iterations to gradually generate a whole face/sketch, as illustrated previously in Fig.~\ref{fig:demo}. In each iteration, backpropagating the loss of $D$ will enforce the photo-reality during the recursive generation. In the case of Fig.~\ref{fig:test_flow}, the backpropagation error is used to adjust the generated sketch $x_\mathcal{S}^k$ as shown in Eq.~\ref{eq:adjust}. Finally, $x_\mathcal{S}^k$ is mapped back to the face domain (Eq.~\ref{eq:S2I}), generating $x_\mathcal{I}^{k+1}$ as an improved version of $x_\mathcal{I}^{k}$ with more details. Repeating this procedure, the large missing area can be filled up gradually. 
	
\begin{figure}[h]
		\centering
		\includegraphics[width=0.8\columnwidth]{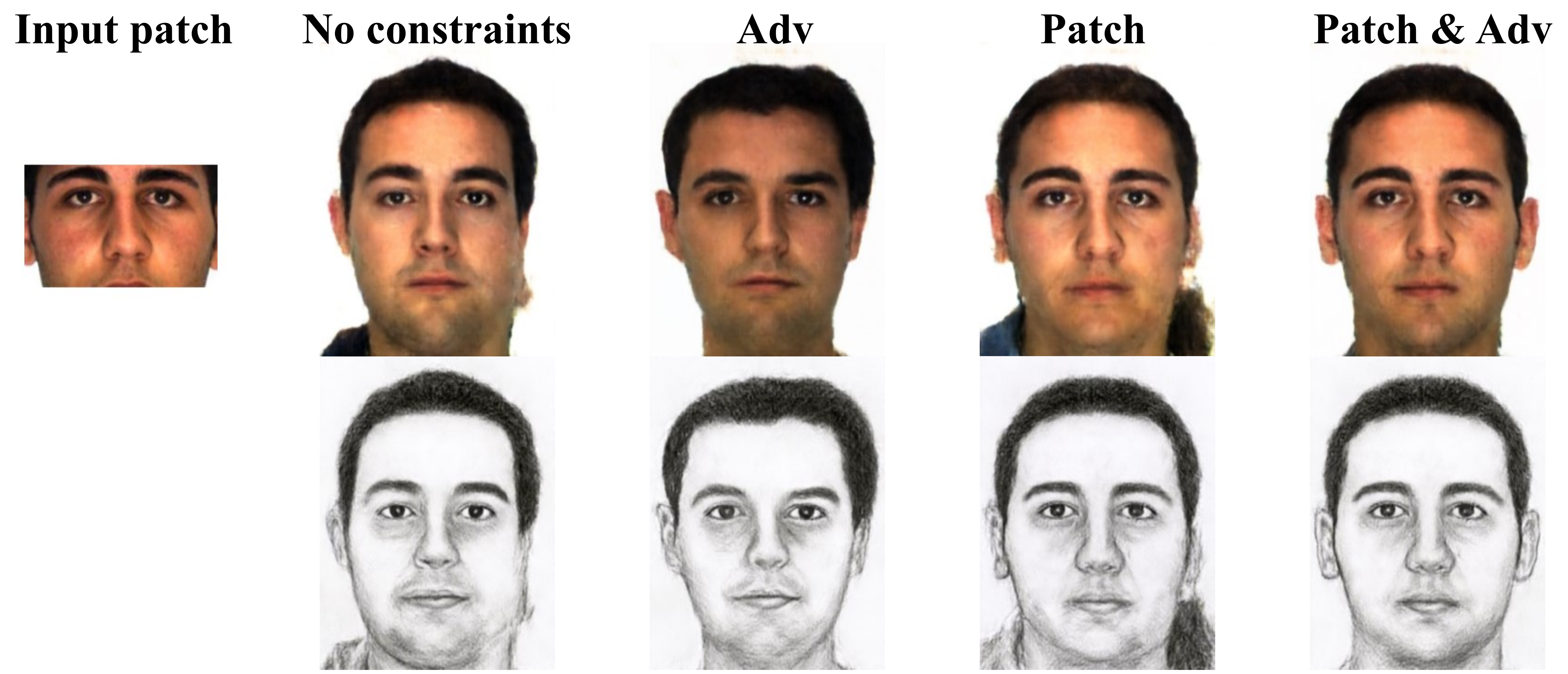}
		\caption{Comparison of generated results with/without the given patch (Patch) and adversarial (Adv) constraints. }
		\label{fig:constraints}
	\end{figure}

	To illustrate the effect of the two constraints, \ie, the given patch and the adversarial constraints, applied during the testing stage,  Fig.~\ref{fig:constraints} shows the generated results with/without the constraints. The given patch and the adversarial constraints are denoted as ``Patch'' and ``Adv'', respectively. It is interesting to observe that the generated face/sketch without ``Patch'' (the second and third columns) cannot preserve the identity of the input patch, and those without ``Adv'' (the second and forth columns) tend to yield unrealistic face/sketch (\eg, the left ear location) or hair style (\eg, the extra hair below the left ear in the fourth column). The results with both constraints obviously outperform the others.
	
\section{Experiment and Results}
	\subsection{Data Collection}
	We collect 1,577 face/sketch pairs from the datasets CUHK~\cite{wang2009face}, CUFSF~\cite{zhang2011coupled}, AR~\cite{martinez2007ar}, FERET~\cite{phillips2000feret}, and IIIT-D~\cite{bhatt2012memetically}. Because the dataset with face/sketch pairs is limited, we train a face to sketch transformation network based on Pix2Pix~\cite{isola2016image} to generate sketches from faces. We collect frontal face images with uniform background and controlled illumination from datasets CFD~\cite{ma2015chicago}, SiblingsDB~\cite{vieira2014detecting}, and PUT~\cite{kasinski2008put}, as well as from searching engines by keywords like ``XXX University faculty profile''. Finally, we obtain 3,126 face/sketch pairs, from which 300 pairs are randomly selected as the  testing dataset.
	
	\subsection{Implementation Details}
	All the face/sketch images are cropped and well-aligned based on the eye locations, and preprocessed to be uniform white background. The transformations $f$ and $F$ are implemented by the Conv-Deconv network. 
	The discriminator $D$ is implemented by the Conv network but adding a fully-connected layer of single output with the sigmoid activation function. In addition, the input layer is modified to be $256^2\times 6$ because the inputs to $D$ are image pairs. Inspired by \cite{isola2016image}, each Conv layer is concatenated to its symmetrically corresponding Deconv layer, thus more details bypass the bottleneck. In the training, we adopt ADAM~\cite{kingma2014adam} ($\alpha=0.0002,\;\beta=0.5$). Because we utilize $D$ to enforce realistic generations during testing, an approximately optimal $D$ is preferred. Therefore, we update $D$ three times for each update of $f$ and $F$. The parameter $\lambda$ in Eq.~\ref{eq:objective} is set to be 100. Details are  shown in supplementary materials.
	After 100 epochs, we could achieve the results as shown in this paper.
	
	During testing, given a small patch from either the face or the sketch domain, it will be transformed recursively as discussed in testing stage. Empirically, the generated images will have most facial features filled quickly at the first five to ten iterations and then tend to converge after 50 iterations. The results shown in this paper are mostly obtained at the 100th iteration.  

	\subsection{Qualitative Evaluation}

	\subsubsection{Face Composite}
	\label{subsubsec:fuse}
\begin{figure}[ht]
		\centering
		\includegraphics[width=1\columnwidth]{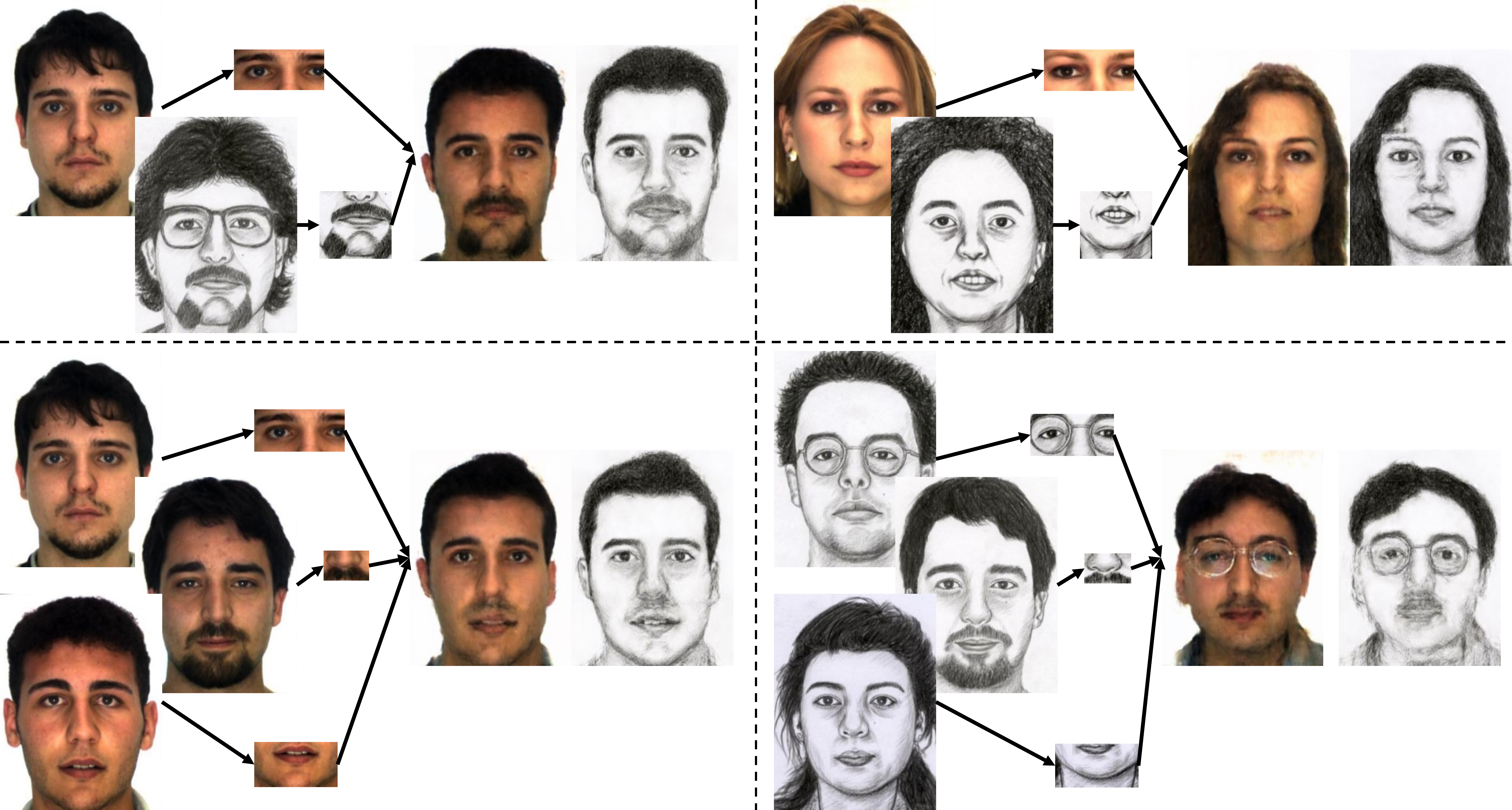}
		\caption{Examples of generated faces/sketches from multiple patches, which are from different people and/or different domains. Four examples are displayed in a 2-by-2 matrix. In each cell, the original faces and sketches are given on the left. The patches are extracted from where indicated by the arrows. The right are generated face/sketch pairs. }
		\label{fig:crossPerson}
	\end{figure}

	We explore the r-BTN to generate consistent and realistic faces from multiple patches that may be from two domains and multiple people. Examples generated from multiple patches are shown in Fig.~\ref{fig:crossPerson}, demonstrating the great versatility of r-BTN. We again observe the strong consistency and fidelity between the generated face/sketch pairs. 
	
	\subsubsection{Face Synthesis from Limited Facial Patches}
	\label{subsubsec:cmp}
\begin{figure*}
	\centering
	\includegraphics[width=1.7\columnwidth]{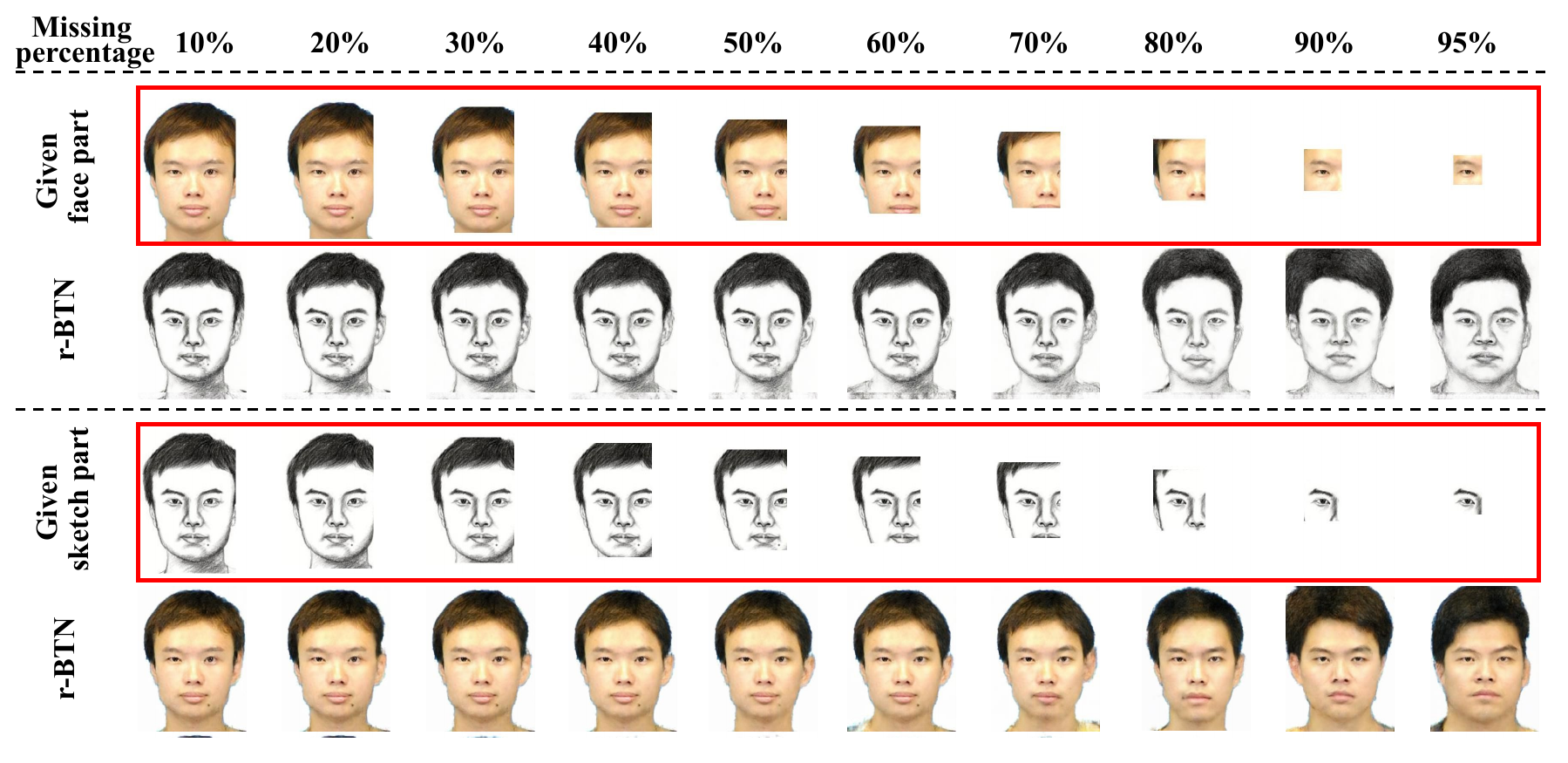}
	\caption{ Comparison in generating faces/sketches from patches with different missing percentage. The red boxes indicate the given face/sketch patches. The rest rows are correspondingly generated sketches/faces.}
	\label{fig:identity_result}
\end{figure*}

\begin{figure*}[t]
		\centering
		\includegraphics[width=1.65\columnwidth]{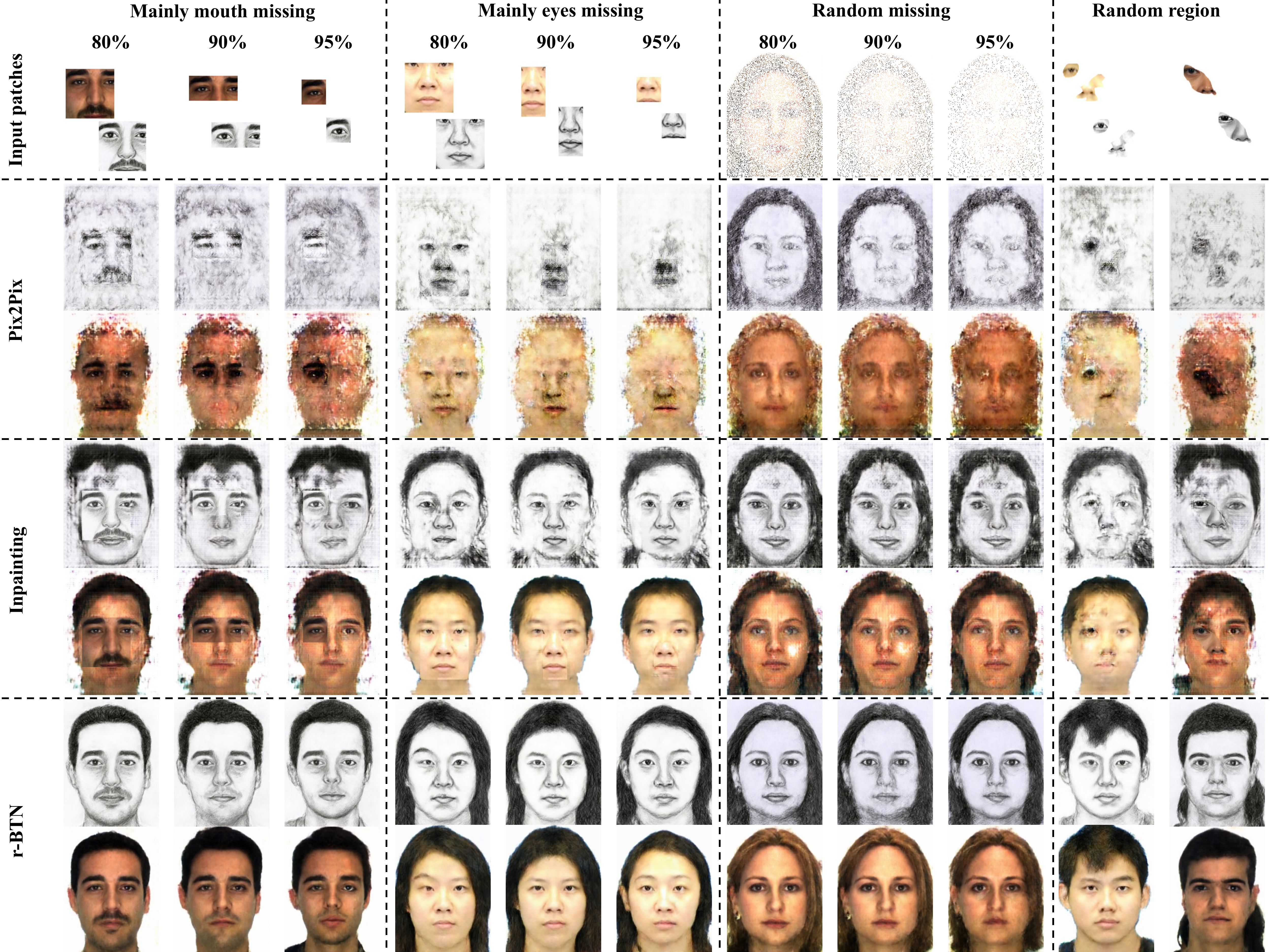}
		\caption{Comparison with other potential methods for filling large missing areas. The first row shows the input patches, and the rest rows display the results from different methods. The percentage indicates missing proportion (missing area over image area). Because Pix2Pix is for domain transfer rather than missing area filling, its results cannot compete with inpainting or r-BTN. We show them here to provide the baseline of domain transfer methods in filling large missing areas.}
		\label{fig:cmp}
	\end{figure*}

   The results generated by proposed r-BTN with respect  to different missing percentage are shown in Fig.~\ref{fig:identity_result}. From the result, it demonstrated that the proposed method cannot preserve the identity when the missing percentage is more than 70\%. This phenomenon is consistent with human cognitive. For human beginnings, if only providing limited information, it is still hard to imagine a unique result. 
	We also compare the proposed r-BTN with Pix2Pix~\cite{isola2016image} and image inpainting~\cite{pathak2016context}. The inpainting method compared in this paper is modified from \cite{pathak2016context} to achieve cross-domain inpainting. Specifically, the inputs are faces/sketches with random mask (20\%$\sim$80\% masked), and the outputs are the whole sketch/face. Pix2Pix and r-BTN are trained with the whole face/sketch pairs. All methods are trained on the same training dataset with the same parameter setting.  The comparison results are shown in Fig.~\ref{fig:cmp}.
	The Pix2Pix and inpainting methods train face-sketch and sketch-face transformation networks independently, so the identity between generated sketches and faces cannot be preserved. 
	For example, comparing the two rows labeled with ``inpainting'', especially the 4th-6th columns, the sketches seem female while the faces appear like male. In addition, the inpainting results present apparent discontinuity between the given patch and the estimated area. On the other hand, the results from r-BTN demonstrate higher fidelity, better consistency to given patches, and better identity preservation. More results are shown in supplementary materials.
		
	\subsection{Quantitative Analysis}
	\subsubsection{Evaluation Metrics}
	\label{subsubsec:metrics}
	To numerically evaluate the quality of generated faces, we design the metric named ``face recognition rate (FRR)". It evaluates whether the generated images present facial elements and geometric structure, \ie, reasonable position of eyebrows, eyes, nose, lips, and chin. We adopt the off-the-shelf face landmark detection method~\cite{kazemi2014one} to detect and localize those facial elements. An unsuccessful detection indicates a failure of face generation. Therefore, FRR is the ratio between the numbers of successfully detected and total generated faces. Fig.~\ref{fig:converge} (left) shows FRR of each method, computed from 300 generated faces using patches with different missing percentages. We observe that when the missing percentage is larger than 50\%,  Pix2Pix fails to generate reasonable faces while inpainting and r-BTN maintain high and similar FRR. However, we recall from Fig.~\ref{fig:cmp} that inpainting results are not photo-realistic as r-BTN although they are both capable of preserving the facial structure.

	\subsubsection{Convergence of Recursive Generation}
	Will the generated faces/sketches converge to a certain point? How many iterations are sufficient to achieve a photo-realistic result? This section mainly answers these two questions. 
	
\begin{figure}[h]
		\centering
        \includegraphics[width=1\columnwidth]{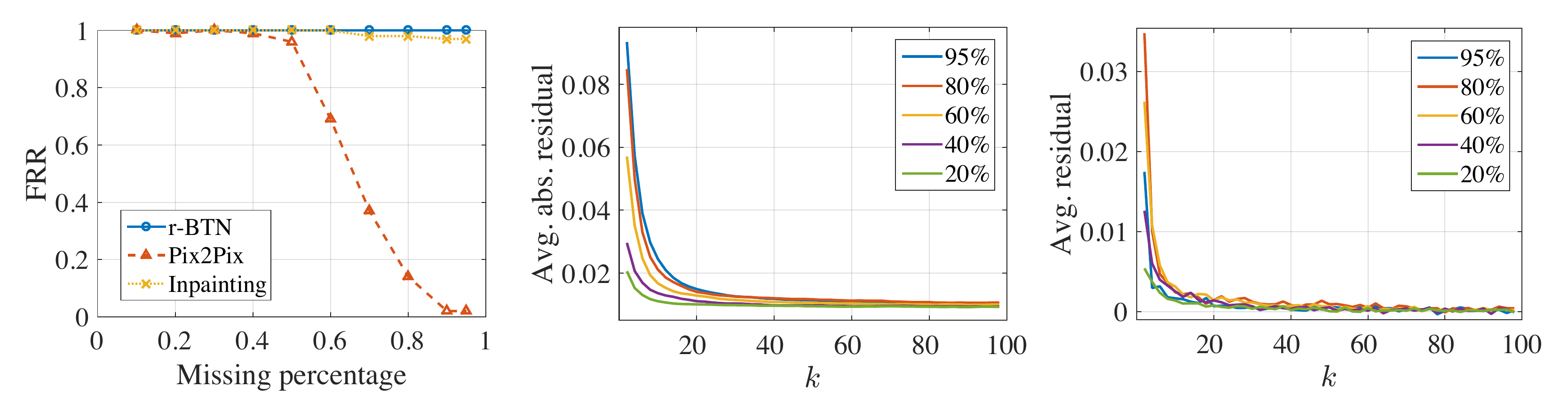}
		\caption{Left:Comparison of different methods on the proposed metrics: FRR. Middle and Right: Convergence evaluation of the proposed r-BTN. Averaged absolute (middle) and average (right) of residual with respect to iteration $k$ are shown at missing percentage of 95\%, 80\%, 60\%, 40\%, and 20\%, respectively.}
		\label{fig:converge}
	\end{figure}

	We first define the residual in the face domain between subsequent iterations as $r^{k+1} = \left( x_\mathcal{I}^{k+1} - x_\mathcal{I}^{k} \right)$, where $x_\mathcal{I}^{k}$ and $x_\mathcal{I}^{k+1}$ denote the $k$th and $k$+1th generated results. The convergence is mainly evaluated by calculating the averaged residual on testing samples (\ie, 300 samples generated  with different missing percentage) with respect to $k$ as shown in Fig.~\ref{fig:converge} (middle). However, the average residual is not sufficient to demonstrate the convergence because some pixels may significantly increase while the other decrease with the same level. In this case, we calculate the averaged absolute residual which illustrate the changing amplitude as shown in Fig.~\ref{fig:converge} (right).

	With more iterations, the averaged residual approaches zero while the averaged absolute residual stabilizes at a small  value. This well demonstrates that the generated faces are stable. In addition, from the experiments (\eg, Fig.~\ref{fig:demo} and~\ref{fig:identity_result}), the generated faces/sketches will not significantly change after 20 iterations. Therefore, we could empirically conclude that the recursive generation will converge to certain face/sketch for a given patch.       
    
    	\subsubsection{Similarity/Diversity Evaluation}
\begin{figure}[h]
		\centering
		\includegraphics[width=0.9\columnwidth]{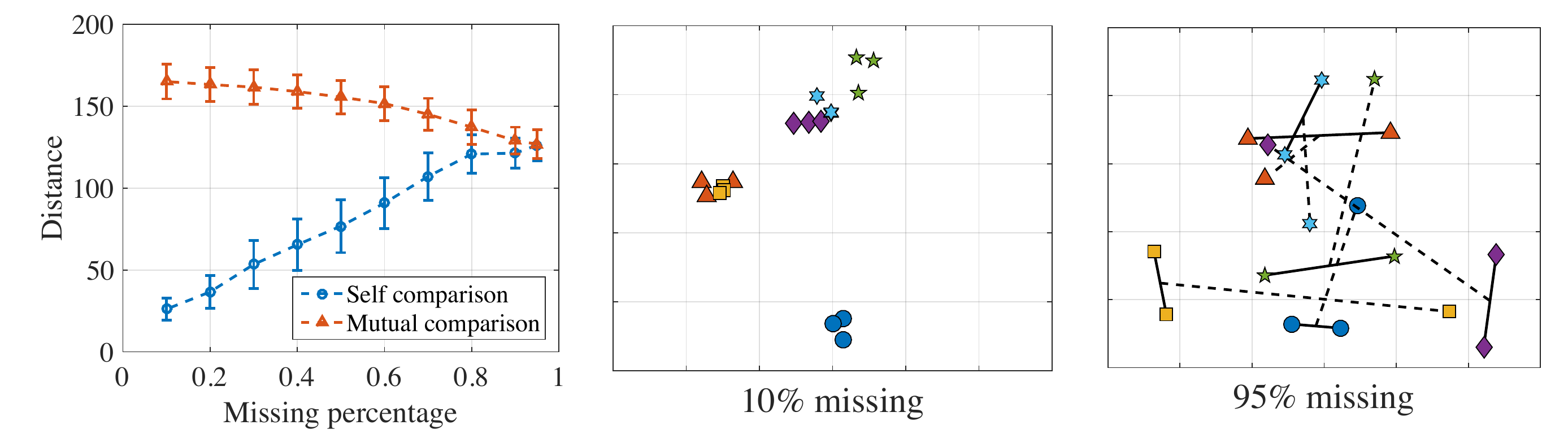}
        \caption{Left: Evaluation of similarity/diversity with increasing missing percentage. The bars indicate corresponding standard deviation. Middle and Right: High-level feature of generated faces at missing percentage of 10\% and 95\%, respectively. There are three same markers for type (person), denoting the generated faces from patches around left eye, right eye, and mouth. Solid lines connect the faces generated from eyes, and the dashed lines connect to the faces generated from mouth.}
		\label{fig:curves}
	\end{figure}

	Intuitively speaking, the generated faces from the patches of the same person should be similar. By contrast, patches from different persons are supposed to yield diverse faces. To verify this property, we collect 50 faces and pick patches of different size around the eyes, the nose, and the mouth. The proposed r-BTN is then applied to generate full faces from those patches. To measure the similarity/diversity between generated faces, we utilize the pre-trained VGG-Face~\cite{parkhi2015deep} model to extract high-level features and compute their Euclidean distance. We perform two comparisons: 1) self comparison (similarity) and 2) mutual comparison (diversity), conducting on faces generated from patches of the same and different persons, respectively.
	Fig.~\ref{fig:curves} (left) shows the averaged distance and standard deviation with respect to missing percentage. The blue circles shows the results of self comparison, and the red triangles denote mutual comparison. 

	With lower missing percentage, \eg, 0.1 to 0.6, the generated faces preserve relatively high intra-class (same person) similarity and inter-class (different persons) diversity. As the missing percentage increases, the two curves eventually intersect, indicating the generated faces from very small patches (\eg, 95\% missing) have lost the identity of the original face. Interestingly, we discover that the generated faces from either the left or right eye of the same person still tend to be more similar as compared to those generated from nose/mouth as illustrated in Fig.~\ref{fig:curves} (right). This discovery is well in line with the quality of different biometrics where studies have shown eyes to carry more valuable cues than nose or mouth in face recognition tasks. This finding, from another perspective, demonstrates the high effectiveness of r-BTN in generating high-fidelity and realistic faces/sketches.

	\section{Discussion and Future works}
	In this paper, we proposed and solved the challenging task of cross-domain face generation with large missing area. A novel recursive generation method by bidirectional transformation networks (r-BTN) was proposed to achieve high-fidelity and consistent face/sketch even with as large as $95\%$ missing area. We demonstrated the effectiveness of r-BTN by comparing to some potential solutions like pix2pix and inpainting. However, r-BTN requires well-aligned faces/sketches. Otherwise, the generated results may not be visually pleasing because the network would fail to localize facial components and thus missing their geometric structure. In the future, we plan to improve the proposed r-BTN from four perspectives: 1) concatenating a face calibration mechanism to r-BTN to battle against the alignment problem, 2) extending this work to be unsupervised  like~\cite{du2013large,taigman2016unsupervised} to alleviate the requirement for paired dataset, 3) generalizing r-BTN as a framework for cross-domain transformation, especially with large missing area, and further evaluating the performance on other datasets~\cite{zhang2018couple}, and 4) adpting this for mobile network application~\cite{li2016deepcham}.

\bibliography{references}
\bibliographystyle{aaai}
\end{document}